\documentclass[pdflatex,sn-basic]{sn-jnl}

\usepackage{graphicx}%
\usepackage{multirow}%
\usepackage{amsmath,amssymb,amsfonts}%
\usepackage{amsthm}%
\usepackage{mathrsfs}%
\usepackage[title]{appendix}%
\usepackage{xcolor}%
\usepackage{textcomp}%
\usepackage{manyfoot}%
\usepackage{booktabs}%
\usepackage{algorithm}%
\usepackage{algorithmicx}%
\usepackage{algpseudocode}%
\usepackage{listings}%
\usepackage{diagbox}
\usepackage{makecell}
\usepackage{subfig}

\theoremstyle{thmstyleone}%
\newtheorem{theorem}{Theorem}%
\newtheorem{proposition}[theorem]{Proposition}%

\theoremstyle{thmstyletwo}%

\theoremstyle{thmstylethree}%

\begin{document}

\title[FedDAF]{FedDAF: Federated Domain Adaptation Using Model Functional Distance}

\author[1]{\fnm{Mrinmay} \sur{Sen}}\email{senmrinmay@alumni.iith.ac.in}
\equalcont{These authors contributed equally to this work.}

\author*[2]{\fnm{Sidhant} \sur{Nair}}\email{sid.nairiitd@gmail.com}
\equalcont{These authors contributed equally to this work.}

\author[1]{\fnm{C Krishna} \sur{Mohan}}\email{ckm@cse.iith.ac.in}

\affil[1]{\orgdiv{Department of Artificial Intelligence}, \orgname{Indian Institute of Technology Hyderabad}, \orgaddress{\city{Hyderabad}, \postcode{502285}, \state{Telangana}, \country{India}}}

\affil*[2]{\orgdiv{Department of Mechanical Engineering}, \orgname{Indian Institute of Technology Delhi}, \orgaddress{\city{New Delhi}, \postcode{110016}, \state{Delhi}, \country{India}}}

\abstract{Federated Domain Adaptation (FDA) improves model performance at a target
client by collaborating with source clients while preserving data privacy. FDA
faces two key challenges: domain shift between source and target data, and lim-
ited labeled data at the target, a common constraint when a new site joins a
federation before it has accumulated its own labeled data, as in clinical deploy-
ments. Most existing methods address domain shift alone,
assuming ample target data; those that also tackle data scarcity still fail to prior-
itize source information according to the target’s specific objective. We propose
FedDAF, which addresses both challenges through similarity-based aggregation
of the global source and target models, using their model functional distance,
computed from the angle between their mean gradient fields on target data and
normalized via a Gompertz function. The global source model itself is formed
using a distance-based weighted average, giving greater weight to source models
closer to the target model. Experiments on real-world datasets show FedDAF
outperforms existing federated learning (FL), personalized FL, and FDA methods in test accuracy.}


\footnotetext[1]{Code is available at \url{https://github.com/sid0nair/FedDAF}}

\maketitle

\section{Introduction}\label{sec:intro}

Centralized training of deep neural networks has shown promising results in terms of model generalization on unseen data. However, this approach often requires a large amount of data, which may not always be available. Collecting sufficient training samples from other sources can be challenging due to financial constraints and privacy concerns. For instance, training a predictive model for breast cancer diagnosis faces significant challenges in obtaining enough annotated data. Manual annotation of medical data requires expert involvement, which can be both costly and time-consuming. Additionally, not all hospitals may have a sufficient volume of patient data, requiring the collection of data from other institutions for centralized training. Sharing data across hospitals, however, raises privacy concerns.

Federated learning (FL) \citep{McMahanaistats2017} presents a promising solution to these challenges by enabling collaborative training of a global model using locally trained models from different sources or clients, without the need to transfer raw data to a central server. This approach helps maintain data privacy. However, despite FL's ability to preserve privacy, the generalization of the global model can be compromised due to domain shift between the training data across different clients \citep{FL_survey_HuangYSWLDY24,Zhao2018}. This issue arises from drift in local models \citep{Karimireddyicml2020,Gaocvf2022} caused by inconsistencies in objectives across clients \citep{Wangneurips2020,Licvpr2021,KairouzMABBBBCC21}. As a result, the aggregated local models (i.e., the global model) may converge to a suboptimal solution, which does not align with the global objective (the average objective across all sources), leading to poor generalization on certain clients \citep{Liiclr2020,Fallahcorr2020,Tancorr2021}. For example, domain divergence in breast cancer data across hospitals can occur due to variations in patient demographics, disease progression, and imaging techniques. Critically, a hospital joining such a federation rarely has a large labeled dataset available from day one, the labeled cases accumulate over time, and any adaptation method deployed at onboarding must work with only the small initial set. This is the deployment constraint our scarcity experiments (Section~\ref{subsec:dataset}, target-label proportions as low as $5\%$) are designed to reflect.

Existing methods address domain divergence through personalized federated learning (PFL) \citep{Fallahcorr2020,Wuieee2022,DinhTN20NeurIPS,li2021ditto,zhang2023fedcr,liu2023feddwa,zhang2020personalized,xu2023personalized}, which aims to customize each local model rather than maintaining a single global model. However, current PFL methods often require substantial amounts of labeled data to train these personalized models, which can be difficult for clients with limited labeled samples \citep{Jiang2024FedGP}. To solve this, federated domain adaptation (FDA) \citep{PengHZS2020fedadDA} has been proposed, where a target client improves its model performance by collaborating with source clients that have sufficient labeled data. However, most of the existing FDA methods \citep{PengHZS2020fedadDA,Feng2021Unsupervised_DA,WuG2021_fedDOMAIN,Zhao2024_more_better,Yao2022_MUTI_TARGET} are proposed to solve the issue of domain shift with the assumption of availability of a large number of unlabeled target samples; tackling both the challenges of domain shift and limited target data has received less attention \citep{Jiang2024FedGP}. Furthermore, existing methods designed to address both issues fail to share relevant information from source clients to the target client in a way that is tailored to the target's specific objectives, which is crucial for adapting the relevant source information based on the target domain.

In this paper, we propose FedDAF, a method designed to address both domain shift and limited target data in Federated Domain Adaptation (FDA). Our approach focuses on increasing the sharing of relevant information from source clients to the target client, based on the target client's specific objective. This is achieved by optimizing the target model at the optimum of the target objective, which is positioned at an optimized distance from the global source objective (the average of all source clients' objectives). To accomplish this, we introduce novel aggregation techniques for the global source model and the target's local model, leveraging their similarity. This similarity is computed using the model functional distance between them, which is determined by the mean gradient fields with respect to the target objective, built using target data. The similarity between two models is effectively measured through this functional distance, even when only limited samples are available. This method allows for the computation of similarity based on the target objective, ensuring that, despite limited sample data, the similarity measure remains reliable \citep{SongXWCS2023modelGIF}. The similarity-based aggregation of the source and target models is designed to increase the relevant information introduced into the target model from the source models, aligned with the target objective. We validate our proposed method through extensive experiments on real-world FDA frameworks, demonstrating that FedDAF outperforms various Federated Learning (FL), Personalized Federated Learning (PFL), and FDA methods in terms of performance.

The main contributions of this paper are as follows:

\begin{itemize}
\item To address the challenges of domain shift and limited target data in federated domain adaptation (FDA), we propose FedDAF. The goal of FedDAF is to increase the transfer of relevant source information to the target model, aligning with the target objective, in order to effectively train an adapted target model.

\item To achieve this, we introduce a novel method for aggregating the global source model and target model based on their similarity, which is quantified using model functional distance. This distance is computed from their mean gradient fields evaluated on the target data, facilitating increased integration of relevant source information into the target model based on the target objective, even when the target dataset is limited.

\item Aggregation of the target model and the global source model, based on the similarity score computed using model functional distance, optimizes the adapted target model (i.e., the aggregated model) at the optimum of the target objective, at a distance from the global objective of the source clients that is designed to increase the transfer of relevant source information into the target model, based on the target's specific objective.
\end{itemize}

The rest of the paper is organized as follows: Section~\ref{prbfor} shows the problem formulation of FDA, Section~\ref{rel_w} describes related works, Section~\ref{propo} describes our proposed method, Section~\ref{exp} shows our experimental setup and results, and Section~\ref{con} concludes the paper.

\section{Problem Formulation of FDA}\label{prbfor}

In this section, we formally define the problem of Federated Domain Adaptation and introduce the notation required to describe it.

\textbf{Notations:} Let $\mathbb{D}^T$ be the dataset of the target client with limited samples. The target objective is defined as
\[
\mathbb{F}^T (\textbf{w}^T; \mathbb{D}^T) = \frac{1}{|\mathbb{D}^T|}\sum_{\psi_i \in \mathbb{D}^T} f^T_i(\textbf{w}^T; \psi_i)
\]
where $f^T_i(\textbf{w}^T; \psi_i)$ is the loss on sample $\psi_i \in \mathbb{D}^T$ and $\textbf{w}^T$ is the local model of the target client.

There are $K$ source clients with overall dataset $\mathbb{D}^S = \mathbb{D}^S_1 \cup \mathbb{D}^S_2 \cup \cdots \cup \mathbb{D}^S_K$, where $\mathbb{D}^S_k$ is the dataset owned by source client $k$. The local objective of source client $k$ is
\[
\mathbb{F}^S_k (\textbf{w}_k^S; \mathbb{D}^S_k) = \frac{1}{|\mathbb{D}^S_k|}\sum_{\psi_i \in \mathbb{D}^S_k} f^S_i(\textbf{w}_k^S; \psi_i)
\]
where $f^S_i(\textbf{w}_k^S; \psi_i)$ is the loss on sample $\psi_i \in \mathbb{D}^S_k$, and $\textbf{w}^S_k$ is the locally trained model of source client $k$. The global source model is given by $\textbf{w}^S = \sum^K_{k=1} p_k \textbf{w}^S_k$, obtained by optimizing the global source objective
\[
\mathbb{F}^S(\textbf{w}^S; \mathbb{D}^S) = \sum_{k = 1}^K p_k \, \mathbb{F}^S_k (\textbf{w}^S; \mathbb{D}^S_k), \quad \text{where } \sum_{k = 1}^K p_k = 1
\]
In federated domain adaptation, the global source model $\textbf{w}^S$ computed by aggregating locally trained source models may not perform well on the target client with limited samples, as the global source optimum is shifted from the target optimum due to domain shift. With this issue, the objective of federated domain adaptation is to aggregate the locally trained model ($\textbf{w}^T$) of the target client and the global source model ($\textbf{w}^S$) of $K$ source clients to compute a new adapted target model ($\textbf{w}$), which can generalize well on the target data. The core idea is to fuse the relevant source information into the target model by using an aggregation weight $\alpha \in [0,1]$ as shown in Eq.~\ref{Eq:1}.

\begin{equation}
    \textbf{w} = \alpha\textbf{w}^S + (1- \alpha)\textbf{w}^T
    \label{Eq:1}
\end{equation}

\section{Related Work}\label{rel_w}

The related works aimed at addressing the model generalization issue for a target client in federated learning, caused by domain shift, can be categorized into two groups: Personalized FL and Federated Domain Adaptation.

\subsection{Personalized FL}\label{subsec:pfl}

Instead of training a single global model, Personalized FL (PFL) algorithms improve the model generalization of each local client by customizing the aggregation for each client based on its requirements, incorporating relevant information from other clients. This addresses the issue of model generalization of the single global model in FL, which is caused by domain shift across clients. Existing PFL methods include pFedMe \citep{DinhTN20NeurIPS}, Ditto \citep{li2021ditto}, FedRep \citep{collins2021exploiting}, FedALA \citep{zhang2023fedala}, FedDWA \citep{liu2023feddwa}, etc. pFedMe adds Moreau envelopes as a regularization term to the local objective function while customizing the local model. Like pFedMe, Ditto trains a personalized model for each client using a proximal term with the local objective. FedRep finds the global feature extractor by aggregating the locally trained feature extractors of all the clients and then fine-tunes the classifier for each local client separately using its local data to create a personalized classifier for this client. FedALA aggregates the global model received from the server with the local model for each client, using separate weights for each parameter. These weights for all the parameters are trained adaptively using a gradient descent optimizer on the local data. FedDWA trains a separate local guidance model alongside the local model by performing one-step adaptation and uses this guidance model on the server to customize the aggregation for each local client.

These existing PFL methods require substantial amounts of labeled data to train personalized models, which can be difficult for clients with limited labeled samples \citep{Jiang2024FedGP}.

\subsection{Federated Domain Adaptation}\label{subsec:fda}

Existing Federated Domain Adaptation (FDA) methods include KD3A \citep{Feng2021Unsupervised_DA}, FADA \citep{PengHZS2020fedadDA}, COPA \citep{WuG2021_fedDOMAIN}, FMTDA \citep{Yao2022_MUTI_TARGET}, FedGP \citep{Jiang2024FedGP}, etc. KD3A applies knowledge distillation on the source models to measure their importance on the target domain and, based on this, calculates the aggregation weights of the source model. FADA employs adversarial techniques to align representations learned across different nodes with the target node's data distribution. COPA uses a domain-invariant feature extractor and domain-specific classifiers. It optimizes local models for each domain, then aggregates feature extractors and classifiers to build a global model. FMTDA introduces a dual adaptation approach that divides the training framework into two parts: local adjustments on client devices and global adaptation on the server. Even though all these FDA methods show promising performance in tackling the domain divergence issue in FL, there is one major challenge with them, i.e., the requirement of a large number of unlabeled target samples in the target domain, which is often impractical \citep{Jiang2024FedGP}.

To address both the issues of limited target samples and domain divergence in FDA, FedGP is proposed. FedGP computes the average of the weighted averages of the target gradient update and the projection of the target gradient update onto each source gradient update, thereby incorporating source information into the target gradient update. For each source gradient update, the aggregation weight is computed by dividing the variance of the target client's intermediate gradient updates by the sum of this variance and the distance between the target and the source gradient updates. Since FedGP calculates the aggregation weight for the source client's gradient update by considering the distance between the target gradient update and the source gradient update, this aggregation does not effectively fuse information from source clients into the target gradient update based on the specific objective of the target client. This is crucial for adapting the source client's domain to the target client's interests.

Aligned with the objective of FedGP, our proposed method is designed to fuse source models with the target model based on the target client's objective, with the aim of increasing relevant information sharing from the source clients to the target client while considering the target's specific objective.

\begin{algorithm}[!h]
   \caption{FedDAF}
   \label{alg:algo1}
\begin{algorithmic}[1]
    \item \textbf{Input:} $N$: FL communication rounds, $\textbf{w}_0^S$: Randomly initialized global model of source clients, $\eta_T$: Target learning rate, $\eta_S$: Source learning rate, $\mu$: Parameter of Gompertz function \newline
    \item \textbf{Return:} $\textbf{w}_N^T$: Final trained target model \newline
    \For{$n=1$ {\bfseries to} $N$}

        \State Server broadcasts global source model $\textbf{w}_{n-1}^S$ to all the source and target clients

        \State \underline{\textbf{In Target client:}}\\
        \If{ $n \geq 2$}
            \State Aggregate $\textbf{w}_{n-1}^T$ and  $\textbf{w}_{n-1}^S$ to compute adapted target model $\textbf{w}_{n}$
        \Else
            \State $\textbf{w}_{n} \leftarrow \textbf{w}_0^S$
        \EndIf
        \State Calculate updated target model $\textbf{w}_{n}^T$ using $\textbf{w}_{n}$ as initial model
        \State Broadcast $\textbf{w}_{n}^T$ to the server

        \State \underline{\textbf{In Source clients:}}\\

        \For{client i $\in \{1, 2, ..., K \}$ \textbf{in parallel}}
            \State Compute updated source model $\textbf{w}_{in}^S$ and broadcast to the server

        \EndFor
        \State \underline{\textbf{In server:}}\\
        \State Receive set of source models $\{\textbf{w}_{in}^S\}$
        \State Aggregate $\{\textbf{w}_{in}^S\}$ using distance-based softmax weighting (Eq.~\ref{Eq:softmax}) to compute updated global source model $\textbf{w}_{n}^S$

    \EndFor
\end{algorithmic}
\end{algorithm}

\section{Proposed Method}\label{propo}

\begin{figure*}[htbp]
    \centering
    \includegraphics[width=120mm]{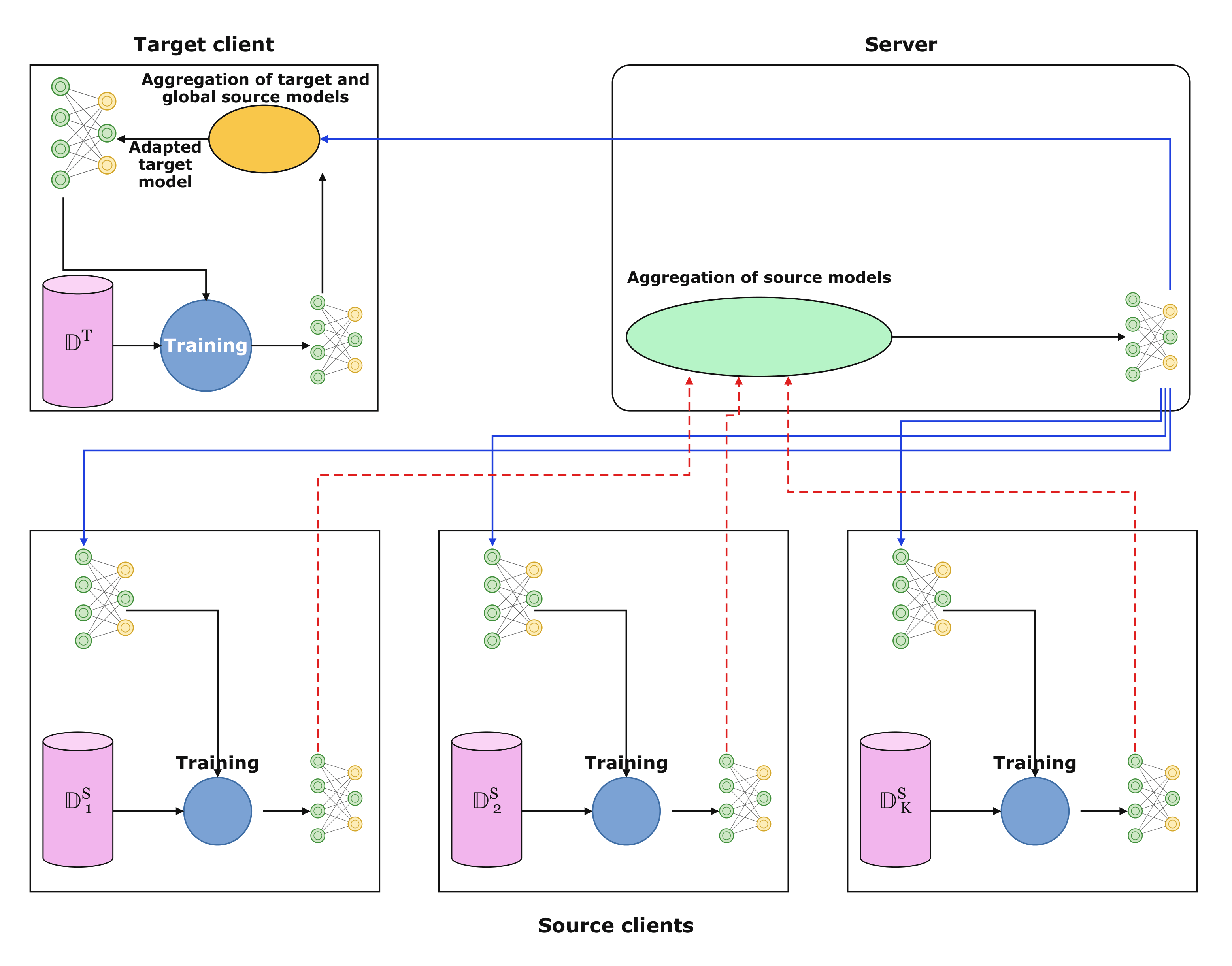}
    \caption{This figure shows the flow diagram of our proposed method. The solid blue arrows indicate the model broadcast from the server to the source and target clients, and the dashed red arrows indicate the transfer of locally trained models from clients back to the server. Each circle denotes a local training operation, and each ellipse denotes a model aggregation operation}
    \label{fig:flow}
\end{figure*}

In this section, we describe our proposed method, FedDAF, designed to improve the target client's model performance under the circumstances of domain shift between the target and source clients, as well as limited target data. The objective behind designing FedDAF is to increase the incorporation of relevant information from the source models to the target model based on the specific objective of the target client to improve the generalization of the target model.

Our proposed method is shown in Algorithm~\ref{alg:algo1} and Figure~\ref{fig:flow}. In FedDAF, in each communication round $n \in \{1, 2, ..., N\}$, the host server first broadcasts the global source model $\textbf{w}_{n-1}^S$ to all the source clients along with the target client. The target client aggregates the global source model $\textbf{w}_{n-1}^S$ and the previous target model $\textbf{w}_{n-1}^T$ based on their similarity, computed using model functional distance through their mean gradient fields derived on the target data. This aggregation results in the formation of the adapted target model $\textbf{w}_{n}$, which is used as the initialization for target-side training. Computation of this adapted target model $\textbf{w}_{n}$ through the model functional distance helps increase relevant information sharing from the source clients to the target client, based on the target objective, by encouraging the optimization of the adapted target model $\mathbf{w}_n$ at the optimum of the target objective $\mathbb{F}^T(\mathbf{w}_{n-1}^T; \mathbb{D}^T)$, which is located at an optimized distance from the optimum of the global source objective $\mathbb{F}^S(\mathbf{w}_{n-1}^S; \mathbb{D}^S)$.

Once the adapted target model is derived, it is then updated using target data and the target optimizer to compute the target model $\textbf{w}_{n}^T$ and broadcast it to the server. In parallel with the computation at the target client, each $i$-th source client separately updates the shared global model from the server and calculates the updated source model $\textbf{w}_{in}^S$ and broadcasts it to the server. The server then aggregates all the locally trained source models $\{\textbf{w}_{in}^S\}$ to compute the global source model $\textbf{w}_{n}^S$.

\subsection{Aggregation of Target and Global Source Models}\label{subsec:agg}

\begin{algorithm}[!h]
   \caption{Compute mean gradient field}
   \label{alg:algo2}
\begin{algorithmic}[1]
    \item \textbf{Input:} $\textbf{w}_{n-1}$: Model, $\mathbb{D}^T$: Target data, $J$: Number of target mini batches \newline

    \For{$ j = 1$ {\bfseries to} $J$}
        \State Compute gradient $ \textbf{g}_{jn}= \frac{\partial \mathbb{F}^T(\textbf{w}_{n-1}; \mathbb{D}^T_{j})}{\partial \textbf{w}_{n-1}}$, where $\mathbb{D}^T_{j} \subseteq \mathbb{D}^T$
    \EndFor

    \State Compute mean gradient field $\textbf{g}_n = \frac{1}{J} \sum_{j = 1}^{J} \textbf{g}_{jn}$
\end{algorithmic}
\end{algorithm}
\begin{algorithm}[!h]
   \caption{Target aggregation}
   \label{alg:algo3}
\begin{algorithmic}[1]
    \item \textbf{Input:} $\textbf{w}_{n-1}^S$: Global source model, $\textbf{w}_{n-1}^T$: Target model, $\mathbb{D}^T$: Target data, $\mu$: Parameter of Gompertz function, $J$: Number of target mini batches \newline
    \State Compute mean gradient field $\textbf{g}_n^S$ of the global source model and mean gradient field $\textbf{g}_n^T$ of the target model on target data using Algorithm~\ref{alg:algo2}
    \State Compute cosine similarity $(sim) =\frac{\textbf{g}_n^T \cdot \textbf{g}_n^S}{||\textbf{g}_n^T|| ||\textbf{g}_n^S||} \in [-1, 1]$
    \State Compute angle $\theta = arccos(sim) \in [0,\pi]$
    \State Normalize $\theta$ using Gompertz function, $\alpha = 1 - e^{-e^{-\mu(\theta - 1)}} \in [0, 1]$, $\mu \in \mathbb{R}$
    \State Aggregate target and global models to find the adapted target model, $\textbf{w}_n= \alpha \textbf{w}_{n-1}^S + (1 - \alpha) \textbf{w}_{n-1}^T$
\end{algorithmic}
\end{algorithm}

\begin{proposition}[Monotonicity of the aggregation weight]\label{prop:monotonicity}
Let $\theta \in [0,\pi]$ be the angle between the mean gradient fields of the target and global source models, and let
\[
\alpha(\theta) = 1 - e^{-e^{-\mu(\theta-1)}}, \quad \mu \in \mathbb{R}
\]
If $\mu > 0$, then $\alpha$ is strictly decreasing in $\theta$, and hence, since $\mathrm{sim} = \cos\theta$ is strictly decreasing on $[0,\pi]$, $\alpha$ is strictly increasing in $\mathrm{sim}$. If $\mu < 0$, the monotonicity is reversed: $\alpha$ is strictly increasing in $\theta$ and strictly decreasing in $\mathrm{sim}$. If $\mu = 0$, $\alpha$ is constant in $\theta$.
\end{proposition}

\begin{proof}
Let $f(\theta) = e^{-\mu(\theta - 1)}$, so $\alpha(\theta) = 1 - e^{-f(\theta)}$. Then
\[
\frac{d\alpha}{d\theta} = f'(\theta)\, e^{-f(\theta)}, \qquad f'(\theta) = -\mu f(\theta)
\]
Substituting,
\[
\frac{d\alpha}{d\theta} = -\mu\, f(\theta)\, e^{-f(\theta)}
\]
Since $f(\theta) > 0$ and $e^{-f(\theta)} > 0$ for all $\theta$, the sign of $\frac{d\alpha}{d\theta}$ is the opposite of the sign of $\mu$. Hence for $\mu > 0$, $\frac{d\alpha}{d\theta} < 0$ for all $\theta \in [0,\pi]$, so $\alpha$ is strictly decreasing in $\theta$; for $\mu < 0$, $\frac{d\alpha}{d\theta} > 0$, so $\alpha$ is strictly increasing in $\theta$; for $\mu = 0$, $\frac{d\alpha}{d\theta} = 0$. Since $\mathrm{sim} = \cos\theta$ is a strictly decreasing bijection from $[0,\pi]$ to $[-1,1]$, the corresponding monotonicity of $\alpha$ in $\mathrm{sim}$ follows immediately.
\end{proof}

In each communication round $n$, the target client receives the global source model $\textbf{w}_{n-1}^S$ from the host server and aggregates it with the previous target model $\textbf{w}_{n-1}^T$ based on the target objective formed by the target data, as shown in Algorithm~\ref{alg:algo3}. To this end, we compute gradient fields of the target model ${\{\textbf{g}_{jn}^T\}}^J_{j =1}$ and gradient fields of the global source model ${\{\textbf{g}_{jn}^S\}}^J_{j =1}$ on target data, here $J$ is the number of mini-batches in the target data. The gradient field for each model is the set of gradients ${\{\textbf{g}_{jn}\}}^J_{j =1}$ computed over different mini-batches on the target data. Once the gradient fields for each of these models are computed, the model functional distance of these two models is measured using the similarity between their gradient fields. To reduce the computation burden associated with computing similarity between these two gradient fields, we compute the mean gradient for each gradient field (as shown in Algorithm~\ref{alg:algo2}), and then compute the cosine similarity ($sim \in [-1, 1]$) between these two mean gradients. Then we compute the angle $\theta \in [0,\pi]$ between them using the inverse of this cosine similarity. We then compute the model functional distance $\alpha \in [0, 1]$ by normalizing this angle $\theta$ using the Gompertz function. This model functional distance is then used as the aggregation weight of the target model and the global source model while computing the adapted target model $\textbf{w}_n$, as shown in Eq.~\ref{Eq:2}.

\begin{equation}
    \textbf{w}_n = \alpha \textbf{w}_{n-1}^S+(1 - \alpha)\textbf{w}_{n-1}^T
    \label{Eq:2}
\end{equation}

\begin{figure*}[ht]
    \centering
    \includegraphics[width=100mm]{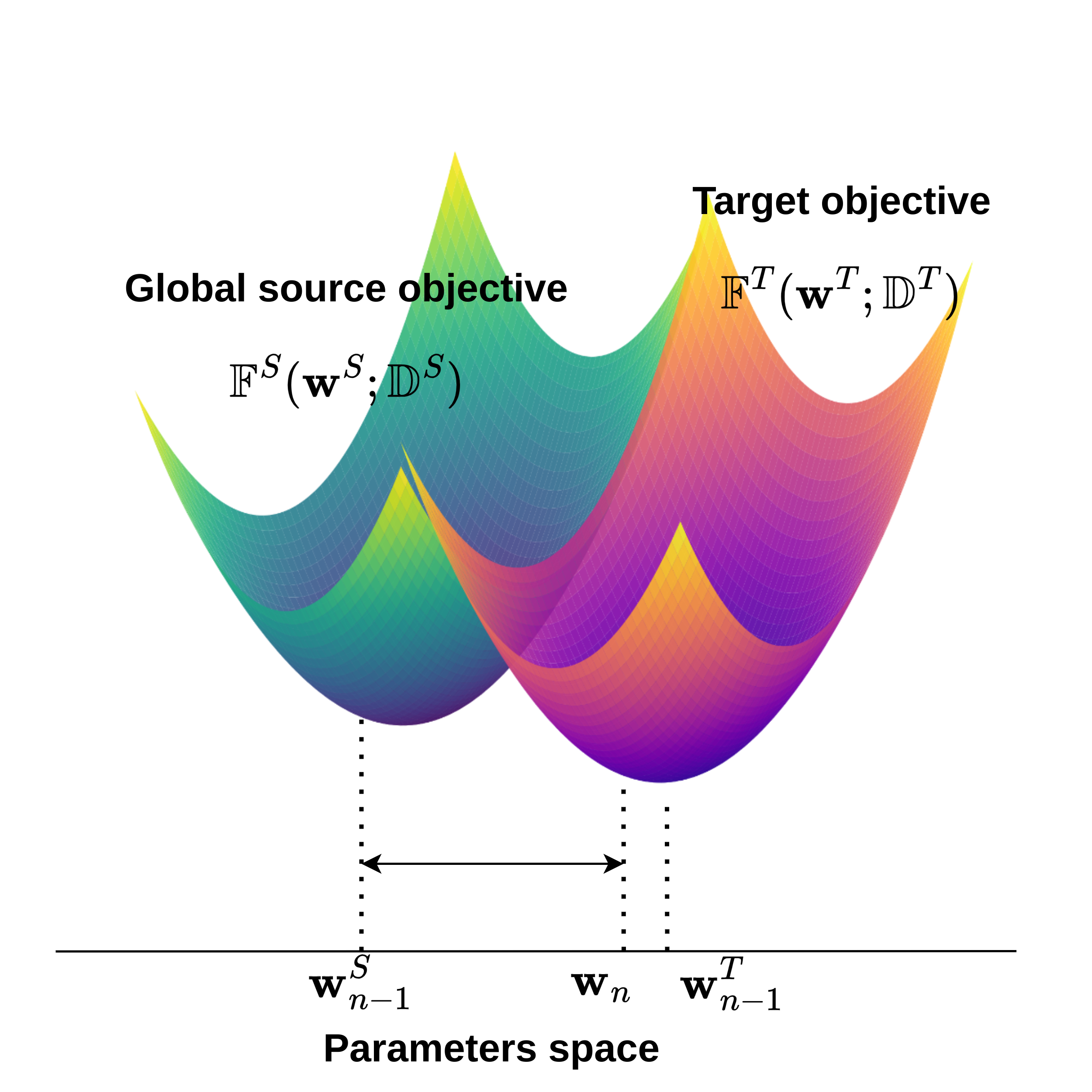}

    \caption{This figure describes the objective of our proposed FDA method. The two peaks represent the global source objective $\mathbb{F}^S(\mathbf{w}^S; \mathbb{D}^S)$ and the target objective $\mathbb{F}^T(\mathbf{w}^T; \mathbb{D}^T)$ in the parameter space. The points $\mathbf{w}_{n-1}^S$ and $\mathbf{w}_{n-1}^T$ denote the optimal global source model and the optimal target model, respectively, and $\mathbf{w}_n$ denotes the adapted target model obtained by aggregating them. The adapted model $\mathbf{w}_n$ lies at an optimized distance from the optimum $\mathbf{w}_{n-1}^S$ of the global source objective, designed to increase the relevant source information fused into the target model after aggregation}

    \label{fig:fig1}
\end{figure*}

To aggregate the global source model and the target model, we use their model functional distance, computed using their model gradient fields. This method aggregates the two models based on the objective defined by the data samples on which the gradient fields are computed, even when only limited samples are available \citep{SongXWCS2023modelGIF}. In our proposed method, we use target samples to compute the model gradient fields, allowing for similarity computation based on the target objective. This ensures that, despite limited target sample data, the similarity measure remains reliable. The optimization of the adapted target model through this aggregation of the target model and the global source model leads to its optimization at the optimum of the target objective, which is located at an optimized distance from the global source objective, as depicted in Figure~\ref{fig:fig1}. This process helps increase relevant information sharing from the source to the target client.

\subsection{Performing Target Training}\label{subsec:target_training}

Once the adapted target model $\textbf{w}_n$ is computed using model functional distance, it is used as the initial model for target training. To prevent the locally updated target model from drifting too far from the global source model, we regularize the target objective with a proximal term, following a FedProx-style formulation:
\[
\tilde{\mathbb{F}}^T(\textbf{w}; \mathbb{D}^T) = \mathbb{F}^T(\textbf{w}; \mathbb{D}^T) + \frac{\lambda}{2}\|\textbf{w} - \textbf{w}_{n-1}^S\|_2^2
\]
where $\lambda \geq 0$ is a proximal coefficient and $\textbf{w}_{n-1}^S$ is the global source model broadcast at the start of round $n$ (Algorithm~\ref{alg:algo1}, line 2). Note that local training is initialized from the adapted target model $\textbf{w}_n$, while the regularizer anchors to $\textbf{w}_{n-1}^S$; the initialization point and the regularization anchor are not the same. We update the target model using $\mathbb{D}^T$ and stochastic gradient descent (SGD) \citep{ketkar2017stochastic} applied to $\tilde{\mathbb{F}}^T$.

\subsection{Performing Source Training}\label{subsec:source_training}

Similar to the target client, each source client $i \in \{1, 2, ..., K\}$ receives the global source model $\textbf{w}_{n-1}^S$ from the server and uses it as the initial model while performing local training with its own local data and optimizer. After local training, all the source clients share their locally trained models $\{\textbf{w}_{in}^S\}$ with the server for creating the updated global source model. At the beginning of the FL training, we randomly initialize the parameters of the global source model $\textbf{w}_0^S$.

\subsection{Aggregation of Source Models}\label{subsec:source_agg}

Once all the locally trained source models $\{\textbf{w}_{in}^S\}$ are received by the server, the server aggregates them using a distance-based weighted average, where each source model's weight depends on its Euclidean distance to the current target model $\textbf{w}_n^T$. Source models that are closer to the target model and therefore more relevant to the target domain are assigned higher aggregation weight. Specifically, we compute
\[
d_i = \|\textbf{w}_{in}^S - \textbf{w}_n^T\|_2, \qquad p_i = \frac{e^{-d_i}}{\sum_{j=1}^K e^{-d_j}}, \qquad i = 1, \ldots, K
\]
and aggregate the source models as
\begin{equation}
    \textbf{w}_n^S = \sum_{i=1}^K p_i\, \textbf{w}_{in}^S
    \label{Eq:softmax}
\end{equation}

\section{Experiments}\label{exp}

To validate our proposed method, we conduct extensive experiments on different real-world datasets in federated domain adaptation frameworks. In this section, we describe the datasets, experimental setup, and analysis of our experimental findings.

\subsection{Dataset}\label{subsec:dataset}

We conduct our experiments on two types of domain shift settings: the first is controlled domain shift, and the second is real-world domain shift.

To create the FDA setting with controlled domain shift, we first randomly divide the entire CIFAR-10 \citep{krizhevsky2009learning} dataset into 50,000 source samples and 10,000 target samples. These 50,000 samples are then divided into 10 source clients using a Dirichlet distribution (concentration parameter = 1), similar to the approach in \citep{yurochkin2019bayesian}. For the target client, we divide the 10,000 samples into $20\%$ for training and the remaining $80\%$ for testing the final adapted target model. To create varying levels of label scarcity in the target client, we use proportions $\in \{0.05, 0.25, 0.5\}$ of the total training samples for target training. To introduce different levels of domain shift between the source and target, we add Gaussian noise with standard deviations $\in \{0.3, 0.6, 0.9\}$ to the target samples.

To create the FDA setting with real domain shift, we use the PACS \citep{LiYSH17PACS}, VLCS \citep{FangXR13VLCS}, and Office Caltech10 \citep{GongSSG12OFCCAL10} datasets, which exhibit real-world domain shifts across all the domains. The PACS dataset contains four different domains: photo (P) with 1670 samples, art painting (A) with 2048 samples, cartoon (C) with 2344 samples, and sketch (S) with 3929 samples. The VLCS dataset also contains four separate domains: VOC2007 (V) with 3376 samples, LabelMe (L) with 2656 samples, Caltech101 (C) with 1415 samples, and SUN09 (S) with 3282 samples. The Office Caltech10 dataset has four different domains: Amazon (A) with 958 samples, Caltech10 (C) with 1123 samples, Dslr (D) with 157 samples, and Webcam (W) with 295 samples. For each dataset, we conduct experiments by considering one target domain and the remaining three as source domains. For the target domain of PACS, we use $2\%$ of the total target data as training samples, and the rest are used for evaluating the adapted target model. Similarly, for VLCS, we use $2\%$ of the target samples for training. For Office Caltech10, we use $2\%$ of target samples when A and C are the target domains, and when D and W are the target domains, we use $20\%$ of the target samples as training samples due to the limited number of available samples.

\subsection{Experimental Setup}\label{subsec:expsetup}
In this section, we describe our experimental setup, including the models used, loss function, performance metrics, compared methods, and implementation details.

\subsubsection{Models and Loss Function}\label{subsubsec:models}
For CIFAR-10 image classification under the FDA setting, we use the ResNet-9 model \citep{he2016deepresnet}\footnotemark[2]. We use the categorical cross-entropy loss for all local training.
\footnotetext[2]{ResNet-9 is a reduced-depth variant of the ResNet architecture family introduced in \citep{he2016deepresnet}, commonly used for lightweight image classification on smaller datasets such as CIFAR-10.}

\subsubsection{Performance Metrics}\label{subsubsec:metrics}
We evaluate the target model using test accuracy, defined as $\frac{\text{Number of true predictions}}{\text{Number of samples}}$ on the target's test dataset, computed in each communication round. We report the best accuracy achieved across all communication rounds.

\subsubsection{Compared Methods}\label{subsubsec:compared}
We compare our proposed method against the following baselines: (i) FedAvg \citep{McMahanaistats2017}; (ii) FedAvg Fine-tune (FedAvg FT), where the target model is initialized from the FedAvg global model and then fine-tuned on the target client's local data; (iii) FedDWA \citep{liu2023feddwa}; (iv) Target only, where the target model is trained using only the target client's local data; and (v) FedGP \citep{Jiang2024FedGP}. Further details on FedDWA and FedGP are provided in Section~\ref{rel_w}.

\subsubsection{Implementation Details}\label{subsubsec:implementation}

For the CIFAR-10 dataset with controlled domain shifts, we use a batch size of source = 64 and batch size of target = 16. For the datasets with real-world domain shifts, we use a batch size of source = 32 and batch size of target = 8. For all the methods, we use a source learning rate $\eta_S \in \{0.01, 0.001\}$ and a target learning rate $\eta_T =\frac{\eta_S}{10}$. For local training, we use the SGD optimizer with one local epoch, and we use $N = 50$ communication rounds. The Gompertz function hyperparameter $\mu$ is set to 5 for our proposed method. The proximal coefficient $\lambda$ in the target training objective is set to $0.001$ for all experiments (Tables~\ref{tab:1}--\ref{tab:3}). For reproducibility of our results, we use seed = 50. For fair comparisons of our proposed method with existing methods, we use the same initialization and settings for all methods on each dataset. While doing a grid search of $\eta_S$, we consider the best-performing model and show it for comparisons. All experiments are performed using a Tesla V100 GPU and PyTorch 1.12.1 with CUDA 10.2.

\subsection{Results}\label{subsec:results}

In this section, we show the results of the experiments along with the empirical analysis, including comparisons with existing methods and a parameter analysis. The comparison results are shown in Tables~\ref{tab:1} and \ref{tab:2}. Table~\ref{tab:1} shows the results of different methods, including existing and proposed methods, on the CIFAR-10 dataset at different levels of data scarcity and domain shift, while Table~\ref{tab:2} depicts the results of three real-world datasets with real-world domain shifts, i.e., PACS, VLCS, and Office Caltech10. The values in both of these tables are obtained by recording the best test accuracy across all the communication rounds achieved by different methods. Table~\ref{tab:3} shows the analysis of the parameter $\mu$ of the Gompertz function, where we record the best test accuracy for different values of $\mu$ on the CIFAR-10 dataset with data scarcity 0.05.

\begin{table*}[t]
\centering
\caption{Best test accuracy on CIFAR-10 under FDA settings with different data scarcity and domain shift levels}
\label{tab:1}
\small
\setlength{\tabcolsep}{4pt}
\newcolumntype{C}{>{\centering\arraybackslash}m{0.9cm}}
\begin{tabular}{l*{9}{C}}
\toprule
& \multicolumn{3}{c}{Scarcity 0.05} 
& \multicolumn{3}{c}{Scarcity 0.25} 
& \multicolumn{3}{c}{Scarcity 0.5} \\
\cmidrule(lr){2-4} \cmidrule(lr){5-7} \cmidrule(lr){8-10}
Method 
& Noise 0.3 & Noise 0.6 & Noise 0.9 
& Noise 0.3 & Noise 0.6 & Noise 0.9 
& Noise 0.3 & Noise 0.6 & Noise 0.9 \\
\midrule
FedAvg      & 14.43 & 14.12 & 13.90 & 14.63 & 13.97 & 13.68 & 14.38 & 14.28 & 14.17 \\
FedAvg FT   & 19.75 & 19.34 & 19.07 & 48.88 & 38.48 & 32.81 & 53.50 & 48.95 & 40.56 \\
FedDWA      & 25.77 & 24.32 & 24.13 & 42.48 & 39.22 & 36.63 & 45.92 & 42.27 & 39.23 \\
FedGP       & 41.35 & 36.55 & 33.21 & 50.91 & 45.60 & 42.10 & 55.06 & 48.66 & 44.23 \\
Target only & 26.75 & 25.59 & 24.75 & 42.06 & 38.73 & 36.33 & 47.32 & 43.60 & 40.78 \\
FedDAF      & \textbf{58.25} & \textbf{43.46} & \textbf{36.27} 
            & \textbf{62.10} & \textbf{49.44} & \textbf{43.13}
            & \textbf{64.43} & \textbf{51.36} & \textbf{44.61} \\
\botrule
\end{tabular}
\end{table*}

\begin{table*}[t]
\centering
\caption{Best test accuracy on real-world domain adaptation datasets}
\label{tab:2}
\small

\begin{tabular}{lcccc}
\toprule
\multicolumn{5}{c}{PACS} \\
\midrule
Method & P & A & C & S \\
\midrule
FedAvg      & 26.51 & 22.01 & 19.32 & 20.64 \\
FedAvg FT   & 42.94 & 30.32 & 43.42 & 36.64 \\
FedDWA      & 44.78 & 29.18 & 45.91 & 48.04 \\
FedGP       & 48.81 & 33.12 & 47.35 & 54.06 \\
Target only & 48.50 & 30.12 & 32.94 & 42.84 \\
FedDAF      & \textbf{60.35} & \textbf{44.67} & \textbf{58.70} & \textbf{58.74} \\
\botrule
\end{tabular}

\vspace{0.4cm}

\begin{tabular}{lcccc}
\toprule
\multicolumn{5}{c}{VLCS} \\
\midrule
Method & V & L & C & S \\
\midrule
FedAvg      & 44.33 & 46.41 & 61.43 & 45.01 \\
FedAvg FT   & 49.05 & 53.54 & 69.35 & 50.01 \\
FedDWA      & 44.48 & 53.86 & 67.12 & 51.85 \\
FedGP       & 50.34 & 56.67 & 69.21 & 53.43 \\
Target only & 45.51 & 52.36 & 68.06 & 46.37 \\
FedDAF      & \textbf{52.70} & \textbf{58.70} & \textbf{71.88} & \textbf{60.18} \\
\botrule
\end{tabular}

\vspace{0.4cm}

\begin{tabular}{lcccc}
\toprule
\multicolumn{5}{c}{Office-Caltech-10} \\
\midrule
Method & A & C & D & W \\
\midrule
FedAvg      & 16.08 & 13.35 & 20.63 & 19.91 \\
FedAvg FT   & 18.10 & 16.62 & 46.03 & 45.76 \\
FedDWA      & 35.78 & 22.07 & 54.76 & 71.18 \\
FedGP       & 50.26 & 30.79 & 77.54 & 78.81 \\
Target only & 40.57 & 23.52 & 63.49 & 75.42 \\
FedDAF      & \textbf{53.25} & \textbf{33.15} & \textbf{83.33} & \textbf{81.78} \\
\botrule
\end{tabular}

\end{table*}

\subsubsection{Comparison with Existing Methods}\label{subsubsec:comparison}

From Table~\ref{tab:1}, it can be observed that when the domain shift is high for any level of data scarcity, the performance of the target model declines. Due to limited data samples, centralized training with target data is not effective. It can also be seen that a higher scarcity of target samples leads to poorer target model performance. Additionally, from this table, it is evident that FedDAF consistently outperforms FedAvg, centralized target training, PFL methods such as FedAvg FT and FedDWA, and the FDA method FedGP, on the FDA setting of the CIFAR-10 dataset, across all levels of data scarcity and domain shifts. From Table~\ref{tab:2}, it can be observed that our proposed method outperforms existing methods in all domains of all datasets with real-world domain shifts, where we consider a very small amount of target training samples to create a higher degree of data scarcity in the target training. Since we use similar settings and the same initialization for all methods on each dataset, we claim that our designed method, FedDAF, is more effective in handling data scarcity in the target client, along with domain shifts between target and source clients, as compared to existing methods.

\begin{table*}[h!]
\centering
\caption{Analysis of Gompertz function's $\mu$ on the performance of FedDAF.}
\newcolumntype{l}{>{\centering\arraybackslash}m{0.9cm}}
\begin{tabular}{*{5}{l}}
\toprule
\multicolumn{2}{c}{Parameter of Gompertz function} & Noise 0.3 & Noise 0.6 & Noise 0.9    \\
\midrule
\multicolumn{2}{c}{$\mu = -10$} &60.93&42.63&34.81   \\
\multicolumn{2}{c}{$\mu = -5$} &60.93&42.51&34.75  \\
\multicolumn{2}{c}{$\mu = -1$} &61.08&42.67&35.37  \\
\multicolumn{2}{c}{$\mu = 0$} &60.73&42.92&35.56  \\
\multicolumn{2}{c}{$\mu = 1$} &\textbf{61.58}&42.85&35.7 \\
\multicolumn{2}{c}{$\mu = 5$} &58.25&\textbf{43.46}&\textbf{36.27} \\
\multicolumn{2}{c}{$\mu = 10$} &31.18&27.66&26.00 \\
\botrule
\end{tabular}
\label{tab:3}
\end{table*}

\subsubsection{Parameter Analysis}\label{subsubsec:paramanalysis}

In this section, we analyze the effect of different values of the Gompertz function's parameter ($\mu$) on the performance of our proposed FedDAF. To this end, we conduct further experiments with $\mu \in \{-10, -5, -1, 0, 1, 5, 10\}$ on the FDA setting of the CIFAR-10 dataset with a data scarcity level of 0.05. The results of these experiments are recorded in Table~\ref{tab:3}. From this table, it can be seen that FedDAF performs better with $\mu = 5$ when the degree of domain shift is comparatively higher. For lower degrees of domain shift, FedDAF performs well with $\mu = 1$. This behavior is consistent with Proposition~\ref{prop:monotonicity}, since $\mu < 0$ reverses the sign of $d\alpha/d\theta$, causing the aggregation weight to increase with gradient-field dissimilarity rather than similarity --- the opposite of the intended mechanism.

\section{Conclusions}\label{con}

In traditional federated domain adaptation (FDA) methods, the domain shift issue in the target client is effectively handled by assuming an ample amount of target data, which may be difficult for targets with a lower number of labeled samples. Both the issues of domain shift and data scarcity in the target client receive limited attention. In this paper, we propose FedDAF, which addresses both issues in FDA by using a model functional distance-based similarity weight for aggregating the target model and the global source model. This approach helps increase the relevant source information incorporated into the target model based on the target client's local objective. The effectiveness of our proposed method is verified and compared with existing methods through real-world FDA experiments.

These results are relevant to the practical constraints of critical-application FL deployments, for instance, a newly onboarded clinical site typically holds only a handful of labeled cases and cannot wait to accumulate a large unlabeled set before benefiting from the federation, which existing FDA methods generally require. We note that this work addresses the domain-shift/data-scarcity problem in isolation from other critical-application requirements: formal differential privacy guarantees, adversarial robustness, and validated deployment in a live clinical system are not evaluated here and are natural directions building on this method.

\backmatter

\section*{Statements and Declarations}

\begingroup
\setlength{\parindent}{0pt}
\setlength{\parskip}{0.6em}

\textbf{Funding} The authors declare that no funds, grants, or other support were received during the preparation of this manuscript.

\textbf{Competing interests} The authors declare that they have no relevant financial or non-financial interests to disclose.

\textbf{Ethics approval and consent to participate} Not applicable.

\textbf{Consent for publication} Not applicable.

\textbf{Data availability} The datasets used in this study, namely CIFAR-10, PACS, VLCS, and Office-Caltech-10, are publicly available from their respective benchmark repositories. CIFAR-10 is available from the official CIFAR dataset page. PACS, VLCS, and Office-Caltech-10 are publicly available benchmark datasets commonly used for domain adaptation and domain generalization. The dataset sources are cited in the manuscript, and the exact download and preprocessing instructions used in this study are provided in the accompanying code repository: \url{https://github.com/sid0nair/FedDAF}.

\textbf{Materials availability} Not applicable.

\textbf{Code availability} The implementation code used for the experiments in this study is publicly available at: \url{https://github.com/sid0nair/FedDAF}.

\textbf{Author contributions} All authors contributed to the conception and design of the study. M.S. and S.N. contributed equally to this work and were involved in method development, experimental implementation, result analysis, and manuscript preparation. C.K.M. supervised the work and provided technical guidance throughout the study. The first draft of the manuscript was written by M.S. and S.N. All authors reviewed and revised previous versions of the manuscript. All authors read and approved the final manuscript.

\endgroup

\bibliography{sn-bibliography}

\end{document}